# Enhancing Neural Network Interpretability Through Conductance-Based Information Plane Analysis

Jaouad DABOUNOU[1*] and Amine BAAZZOUZ[2]

[1*] Professor, Hassan First University of Settat, Faculté Sciences et Techniques, Department of Mathematics Informatics and Engineering Science, Settat, Morocco

[2] Ph.D. student, Hassan First University of Settat, Faculté Sciences et Techniques, Department of Mathematics Informatics and Engineering Science, Settat, Morocco

*Corresponding author E-mail: jaouad.dabounou@uhp.ac.ma;
Contributing author E-mail: amine.baazzoz@uhp.ac.ma

**Abstract**

The Information Plane is a conceptual framework used to analyze the flow of information in neural networks, but traditional methods based on activations may not fully capture the dynamics of information processing. This paper introduces a new approach that uses layer conductance, a measure of sensitivity to input features, to enhance the Information Plane analysis. By incorporating gradient-based contributions, we provide a more precise characterization of information dynamics within the network. The proposed conductance-based Information Plane and a new Information Transformation Efficiency (ITE) metric are evaluated on pretrained ResNet50 and VGG16 models using the ImageNet dataset. Our results demonstrate the ability to identify critical hidden layers that contribute significantly to model performance and interpretability, giving insights into information compression, preservation, and utilization across layers. The conductance-based approach offers a granular perspective on feature attribution, enhancing our understanding of the decision-making processes within neural networks. Furthermore, our empirical findings challenge certain theoretical predictions of the Information Bottleneck theory, highlighting the complexities of information dynamics in real-world data scenarios. The proposed method not only advances our understanding of information dynamics in neural networks but also has the potential to significantly impact the broader field of Artificial Intelligence by enabling the development of more interpretable, efficient, and robust models.

## 1. Introduction

Understanding how information flows through neural networks during training or inference is crucial for interpreting their behavior and improving their design. By analyzing the dynamics of information transfer, we can identify which layers are most influential in learning and decision-making processes. This knowledge helps in diagnosing issues like overfitting, underfitting, and vanishing gradients. Additionally, insights gained from information flow analysis can guide the design of more efficient and robust architectures, ultimately leading to models that generalize better to unseen data and perform more effectively across diverse tasks. This understanding reduces the gap between model performance and interpretability, enhancing both research and practical applications.

The Information Plane, introduced by Tishby et al. (2000), is a groundbreaking framework that leverages information theory to analyze the flow of information through neural networks. The core idea behind this approach is to represent the learning dynamics of a neural network by examining the mutual information between different layers and the input/output data. Mutual information quantifies the amount of information one random variable contains about another, providing a measure of dependence or correlation between variables (Cover, 2006). Subsequent work by Shwartz-Ziv and Tishby (Shwartz-Ziv, 2017) applied this framework to deep neural networks, proposing that the learning process in these networks can be characterized by two distinct phases: an initial fitting phase followed by a compression phase. This perspective has motivated significant interest and debate in the machine learning community, leading to further investigations and refinements of the information bottleneck theory in deep learning (Saxe et al., 2019).

Traditional approaches to visualizing and understanding neural networks often involve examining the outputs of various layers to understand what features each layer detects (Zeiler, 2014; Simonyan, 2014). For instance, early layers might detect edges, while deeper layers capture complex patterns. While these techniques offer insights into model behavior and decision-making processes, they do not provide a clear quantitative measure of how information is processed and transformed within the network layers (Goldfeld et al., 2019). Furthermore, while Information Bottleneck and bifurcation points offer valuable insights during the training phase, they lack a systematic way to quantify the information dynamics within the neural network during inference. These methods primarily highlight important features or activations and how information is compressed, but do not measure the real flow of information through the network. Without a quantitative assessment, it is challenging to understand the overall dynamics and efficiency of information processing within the model.

Recent work by Amjad and Geiger (2020) has explored the application of information bottleneck principles to representation learning in neural networks, providing new perspectives on how information theory can be applied to understand and optimize deep learning models. Additionally, studies like those by Yu and Principe (2019) have demonstrated the utility of information-theoretic concepts in analyzing specific neural network architectures, such as autoencoders.

The present research aims to characterize information dynamics by using layer conductance to highlight the information flow inside neural networks through the information plane. Layer Conductance, a concept derived from the field of explainable AI, is used to understand and interpret the contributions of input features to the activations within specific layers of a neural network. Dhamdhere et al. (2019) introduced a related concept of "neuron conductance" to quantify the importance of individual neurons, which we extend to the layer level in our analysis.

Conductance provides a detailed view of how each input feature influences the neurons in a given layer, allowing researchers to gain insights into the internal workings of the network. Layer conductance is related to the network dynamics because it captures the gradient-based contributions of the input features to the activations of the layer, which measures the sensitivity of a layer's activations for the input features. It can be calculated using the derivative of the activation function concerning the inputs:

$$C_l = \frac{\partial T_l}{\partial X} \cdot X$$

The theoretical framework proposed by Tishby et al. (2000), emphasizes the idea that deep neural networks can be understood through the lens of information theory, particularly the information bottleneck method. However, there are some considerations regarding this framework, especially when real-world data are considered. Empirical data often show that information paths in neural networks could not be monotonic. This contradicts the theoretical prediction that mutual information between $T_i$ and $Y$ (or $X$) should decrease monotonically across layers. The concept of a unique information path in the information plane relies on idealized conditions. Real-world data and network training dynamics can lead to deviations from the ideal path, drawn according to Data Processing Inequality:

$I(X;Y) > I(X;T_1) > I(X;T_2) > \ldots > I(X;\hat{Y})$.

## 2. Theoretical Foundations

In the context of neural networks, the Information Plane plots mutual information in two dimensions: $I(T;X)$ on the x-axis, representing the mutual information between the hidden layer $T$ and the input $X$, and $I(T;Y)$ on the y-axis, representing the mutual information between the hidden layer $T$ and the output $Y$. This visualization allows to observe how information is transformed as it propagates through the layers of the network.

By examining these plots, it is possible to identify phases of learning, such as fitting and compression. During the fitting phase, the network captures relevant information from the input. In the compression phase, the network discards irrelevant information while retaining essential features (Shwartz-Ziv, 2017). This analysis provides deeper insights into the network's behavior during training, revealing how different layers contribute to learning and generalization, and helping to diagnose and mitigate issues such as overfitting or inefficient information use.

### 2.1 Mutual Information:

Mutual Information is defined as:

$$I(X;Y) = \sum_{x \in X} \sum_{y \in Y} p(x,y) \log \frac{p(x,y)}{p(x)p(y)}$$

where $X$ and $Y$ are two random variables, $p(x,y)$ is the joint probability distribution of $X$ and $Y$, and $p(x)$ and $p(y)$ are the marginal distributions (Cover, 2012).

The Information Plane plots two key mutual information quantities:

- $I(X;T)$: Mutual Information between the input $X$ and the hidden representation $T$ (a layer in the neural network).
- $I(T;Y)$: Mutual Information between the hidden representation $T$ and the output $Y$.

These coordinates allow us to visualize how much information each layer retains about the input and how much useful information it has about the output (Tishby, 2015).

### 2.2 Information Plane Dynamics

During the training process, neural network layers evolve in the Information Plane. The typical dynamics observed are:

**Compression Phase**: Early in training, layers retain a lot of information about the input ($I(X;T)$ is high). Over time, layers discard irrelevant information, leading to a reduction in $I(X;T)$, known as the compression phase (Shwartz-Ziv, 2017).

**Fitting Phase**: Simultaneously, the mutual information between the hidden representation and the output $I(T;Y)$ increases, indicating that the network is learning useful features for the task. Layers aim to maximize $I(T;Y)$ while minimizing $I(X;T)$, achieving a balance between retaining relevant information and discarding noise (Achille, 2018).

The Information Plane dynamics are closely related to the Information Bottleneck (IB) principle, which aims to find a compressed representation $T$ of the input $X$ that retains as much relevant information as possible about the output $Y$. The objective is to optimize the following trade-off:

$$\min_{T}[I(X;T) - \beta I(T;Y)]$$

where $\beta$ is a trade-off parameter that controls the balance between compression $I(X;T)$ and prediction accuracy $I(T;Y)$ (Tishby et al., 2000).

### 2.3 Concept of Layer Conductance

Layer Conductance is a concept derived from the field of explainable AI and is used to understand and interpret the contributions of input features to the activations within specific layers of a neural network (Montavon et al. 2018; Zhang, 2020). Conductance provides a view of how each input feature influences the neurons in a given layer, allowing researchers to gain insights into the internal workings of the network. Layer conductance measures the sensitivity of the activations in a neural network layer with respect to the input features.

There are several ways to formulate and compute layer conductance. The following are some of the primary formulations used in practice, and will be adopted in this work:

**a. Gradient-based Conductance**

Gradient-based conductance measures the derivative of the layer's activations with respect to the input features. This formulation is based on the gradient of the activation function (Sundararajan, 2017):

$$C_l = \frac{\partial T_l}{\partial X} \cdot X$$

where:

- $T_l$ represents the activations at layer $l$
- $X$ is the input to the neural network.
- $\frac{\partial T_l}{\partial X}$ is the gradient of the activations with respect to the input.

**b. Integrated Gradients**

Integrated Gradients is a method that accumulates the gradients of the output with respect to the input along a straight-line path from a baseline (typically zero) to the input (Sundararajan, 2017; Shrikumar, 2017; Ancona, 2019):

$$IG_l = (X - X') \cdot \int_{\alpha}^{1} \frac{\partial T_l(X' + \alpha(X - X'))}{\partial X} \, d\alpha$$

where:

- $X'$ is the baseline input.
- $X$ is the actual input.
- $\frac{\partial T_l}{\partial X}$ is the gradient of the activations at layer $l$ with respect to the input $X$.

The term
$$\frac{\partial T_l(X' + \alpha(X - X'))}{\partial X}$$
represents the gradient of the activations at layer $l$ with respect to the input $X$ when the input is $X' + \alpha(X - X')$. Here, $\alpha$ varies from 0 to 1, tracing a straight-line path from the baseline input $X'$ to the actual input $X$.

This method provides a more robust measure of conductance by considering the path integral of gradients.

**c. Interpretation of Layer conductance**

The layer conductance formulation:
$$C_l = \frac{\partial T_l}{\partial X} \cdot X$$
can be expanded as follows:
$$C_l = \sum_{i=1,n} \frac{\partial T_l}{\partial X_i} \cdot X_i$$
where $n$ is the number of input features. This summation indicates that conductance is a weighted sum of the contributions of each input feature $X_i$ to the activations $T_l$.

The conductance $C_l$ not only measures the sensitivity of the activations to input changes but also incorporates the actual values of the input features. The integration of $X$ into the conductance calculation highlights the dynamic nature of information flow and gives a good view of how information is transformed and preserved within the network (Dhamdhere, 2019). This dynamic aspect is crucial for understanding how the network adapts to different inputs and how it prioritizes certain features over others during the information processing stages.

The activations $T_l$ can be considered as a function of the previous layer's outputs, we can apply the chain rule to express the conductance in terms of the previous layer:
$$C_l = \frac{\partial T_l}{\partial X} \cdot X = \frac{\partial T_l}{\partial T_{l-1}} \cdot \frac{\partial T_{l-1}}{\partial X} \cdot X = \frac{\partial T_l}{\partial T_{l-1}} \cdot \frac{\partial T_{l-1}}{\partial T_{l-2}} \cdots \frac{\partial T_1}{\partial X} \cdot X$$

This expression shows how the conductance of layer $l$ depends not only on the input $X$ but also on the activations of the previous layers $T_{l-1}, \cdots, T_1$. It highlights the propagation of information through the network.

## 2.4 Information Plane with Layer Conductance

The Conductance-based Information Plane uses coordinates: $I(X; C)$ and $I(C; Y)$, where $C$ is the conductance at a specific layer.

In this new formulation of the Information Plane, we use $C$ which represents the conductance at a specific layer of the neural network, rather than the activations $T$. The coordinates are defined as follows:

- $I(X; C)$: Mutual Information between the input $X$ and the conductance $C$ at a specific layer.
- $I(C; Y)$: Mutual Information between the conductance $C$ and the output $Y$.

**Interpretation of $I(C;Y)$:**

$I(C;Y)$, the mutual information between layer conductance (C) and the output $Y$, can be seen as a measure of:

- **Relevance:** It quantifies how much of the information captured by the layer's conductance is directly relevant to the final prediction. A high $I(C;Y)$ indicates that the layer's influence is highly informative for the task.
- **Predictive Power:** It reflects the layer's contribution to the overall predictive power of the model. Layers with high $I(C;Y)$ are essential for making accurate predictions.
- **Feature Importance:** It can be used as a proxy for feature importance. If a layer has high $I(C;Y)$, it suggests that the features it captures are important for the task.

**Interpretation of $I(X;C)$**

- **Input Representation**: $I(X;C)$ quantifies how well the conductance $C$ represents the input $X$. This helps in understanding the quality of the internal representations learned by the network.
- **Feature Extraction**: By analyzing $I(X;C)$ across different layers, we can understand how the network extracts and transforms features from the input data.
- **Latent Concepts**: $I(X;C)$ can reveal latent concepts learned by the network, as it measures the relationship between the input and the conductance of a layer.

**Combined Insights**

By combining $I(C;Y)$ and $I(X;C)$, we can gain a comprehensive understanding of the information flow within the network:

- **Redundancy and Compression**: $I(C;Y)$ and $I(X;C)$ can reveal whether the layer retains redundant information or if it effectively acting to compress the input information relevant to the output.
- **Information Flow Dynamics**: The combined analysis of $I(C;Y)$ and $I(X;C)$ can reveal the dynamics of information flow within the network, showing how information is transformed and utilized by each layer.
- **Layer Importance**: By comparing $I(C;Y)$ and $I(X;C)$ across different layers, we can identify which layers are most important for both representing the input and contributing to the final prediction.
- **Interpretability**: The mutual information metrics provide a good interpretable measure of the role of each layer in the network's information flux dynamics and consequently of its decision-making process, making it easier to understand and explain the network's functioning.

## 2.5 Mathematical Foundations of the Conductance-based Information Plan

The conductance of layer $l$, denoted as $C_l$, can be defined as:

$$C_l = \frac{\partial T_l}{\partial X} \cdot X = J_l \cdot X$$

where $J_l$ is the Jacobian matrix of the activations $T_l$ with respect to the input $X$.

$C_l$ is a deterministic function of $X$, the mutual information. $I(X; C_l)$ can be expressed as:

$$I(X; C_l) = H(C_l) - H(C_l \mid X) = H(C_l)$$

where $H(C_l)$ is the entropy of the conductance $C_l$.

If the input $X$ follows a multivariate Gaussian distribution $\mathcal{N}(\mu, \Sigma)$, then the conductance $C_l$ also follows a multivariate Gaussian distribution:

$$C_l \sim \mathcal{N}(J_l \cdot \mu, J_l \cdot \Sigma \cdot J_l^T)$$

The entropy of a multivariate Gaussian distribution is given by:

$$H(C_l) = \frac{1}{2} \log((2\pi e)^d \cdot \det(J_l \cdot \Sigma \cdot J_l^T))$$

Thus, the mutual information $I(X; C_l)$ is:

$$I(X; C_l) = \frac{1}{2} \log((2\pi e)^d \cdot \det(J_l \cdot \Sigma \cdot J_l^T))$$

Here, $\det(J_l \cdot \Sigma \cdot J_l^T)$ captures the weighted sensitivity of the layer $l$ to variations in the input $X$, where the weighting is provided by the input covariance $\Sigma$. This quantity represents a form of "information content" that accounts for the correlations and dependencies among the input dimensions.

The mutual information $I(C_l; Y)$ is defined as:

$$I(C_l; Y) = H(Y) - H(Y \mid C_l)$$

where $H(Y)$ is the entropy of the expected output $Y$, which is constant for a given problem. It represents the uncertainty in the output labels before any information from the conductance is considered. For a classification problem with $K$ classes and assuming $Y$ is uniformly distributed, $H(Y)$ is constant and can be calculated as:
$$H(Y) = \log K$$

$H(Y \mid C_l)$ is the conditional entropy of $Y$ given the conductance $C_l$. Given that $Y$ is represented by one-hot encoding, where only one element is 1 (representing the correct class) and the others are 0, the conditional entropy $H(Y \mid C_l)$ can be approximated based on the assumption that $Y$ is conditionally independent given $C_l$:

$$H(Y \mid C_l) = -E_{Cl} \left[ \sum_y P(Y = y \mid C_l) \log P(Y = y \mid C_l) \right]$$

Since $Y$ is a one-hot encoded vector, where only one element is 1 (representing the correct class) and the others are 0, the entropy simplifies significantly:

$$H(Y \mid C_l) = -\frac{1}{N} \sum_{j=1,N} \sum_{i=1,K} P(Y = i \mid C_l) \log P(Y = i \mid C_l)$$

The conditional entropy is estimated empirically over a dataset with $N$ samples. It averages the conditional entropy across all samples, considering the distribution of $C_l$ across the dataset.

High $I(C_l; Y)$ indicates that the conductance $C_l$ captures significant information about the output $Y$, reflecting an efficient information flow and dynamic processing within layer $l$. This implies that the layer effectively extracts and utilizes features critical for accurate predictions, leading to low conditional entropy $H(Y \mid C_l)$. The reduced uncertainty when $C_l$ is known highlights the layer's role in maintaining robust neural dynamics and optimizing the overall network performance.

Conversely, low $I(C_l; Y)$ suggests that the conductance $C_l$ conveys little information about the output $Y$, indicating inefficient information flow and suboptimal neural dynamics in layer $l$. This results in high conditional entropy $H(Y \mid C_l)$, meaning significant uncertainty about the output persists even with known conductance. Such a scenario suggests that the layer fails to effectively process and transmit relevant features, pointing to potential areas for network improvement and refinement in capturing and utilizing dynamic information.

## 2.6 Information Transformation Efficiency

In our explorations to analyze the network's information dynamic, we propose to quantify the efficiency of information transformation by a layer. It takes into account several critical aspects: information compression, preservation of relevant information, and increase in useful information for the prediction task. Here is a detailed explanation of each component and the overall expression:

- **Information Compression ($C_l$):**

$$C_l = \max\left(0, \frac{-\Delta H_l}{H_{l-1}}\right)$$

  o $H_l$ and $H_{l-1}$ are the entropy of activations of layers $l$ and $l-1$.
  
  o $\Delta H_l = (H_l - H_{l-1})$ is the entropy variation between layers $l$ and $l-1$.

$C_l$ is positive when $\Delta H_l$ is negative (compression). $C_l$ is normalized by the entropy of the previous layer to make it comparable between layers.

This term captures a layer's ability to compress information. Efficient compression (reduced entropy) is often desirable as it can mean that redundant information is eliminated, thus concentrating relevant information. For each layer $l$, the entropy of activations is calculated for a set of inputs.

- **Preservation of Relevant Information ($P_l$):**

$$P_l = \max\left(\frac{I(T_l; T_{l+1})}{\min(H_l, H_{l+1})}\right)$$

$I(T_l; T_{l+1})$ is the mutual information between the activations of layers $l$ and $l-1$.

   $P_l$ measures the proportion of information preserved relative to the available information, normalized to be bounded between 0 and 1. This term evaluates how much relevant information is preserved across layers. The normalization facilitates comparisons.

- **Increase in Useful Information ($U_l$):**

$$U_l = \max_{X_i} \left( \frac{I(T_l\,;Y) - I(T_{l-1}\,;Y)}{H(Y) - I(T_{l-1}\,;Y)} \right)$$

- o   $I(T_l\,;Y)$ is the mutual information between the activations of layer $l$ and the output labels $Y$.
- o   $H(Y)$ est l'entropie des labels de sortie.

$U_l$ measures the relative increase in useful information for the task, normalized by the maximum amount of information that could be gained. It compares the mutual information between a layer's activations and the output labels with that of the previous layer, normalized by the label entropy.

- **Global Efficiency** ($E_l$):

$$E_l = \alpha\, C_l + \beta\, P_l + \gamma\, U_l$$

- o   α, β and γ are adjustable weighting coefficients according to the relative importance of each component...

This combination of entropic variation and information flow can be a good entity to characterize information dynamics. Entropic variation captures changes in the distribution of activations, while information flow measures how information propagates and transforms from one layer to another.

The notion of relevance and efficiency for a neural network layer is not a completely new idea, but it is not always formulated exactly as we have proposed it in the scientific literature.

## 3. Experiments and Discussion

In this section, we present an empirical evaluation of our proposed approach using pretrained models, specifically VGG16 and ResNet50, on the ImageNet dataset. Our proposed Information Transformation Efficiency (ITE) metric and the conductance-based information plane provide a good understanding of the information dynamics in neural networks. The Experiments demonstrate the significant contributions of our proposed approach to understanding information dynamics across neural network layers. First, the Information Transformation Efficiency metric offers a preliminary analysis of information compression, preservation, and utility across different layers. Additionally, the conductance-based information plane offers a complementary perspective by focusing on the gradient-based contributions of input features to layer activations.

**a. Information Transformation Efficiency (ITE) Analysis**

Fig. 1 and Fig. 2 show for VGG16 and ResNet50 a sort of alternance between high and low Information Compression. At the opposite, the preservation of relevant information is locally high for specific layers. The Usefulness can be considered as following the same shape as Compression.

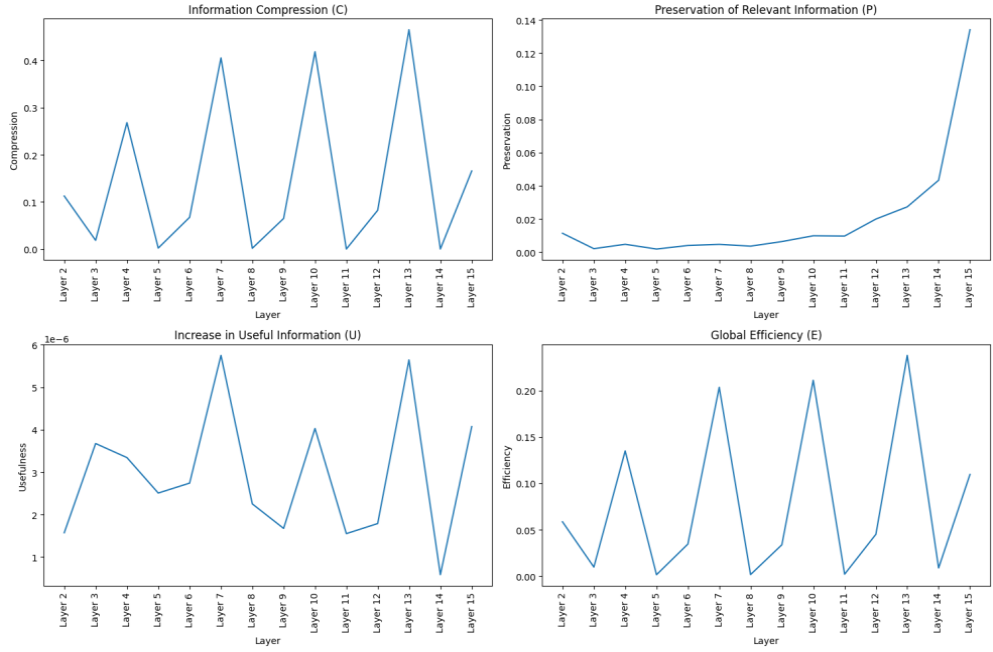

**Fig. 1:** Information Transformation Efficiency for VGG16 on ImageNet

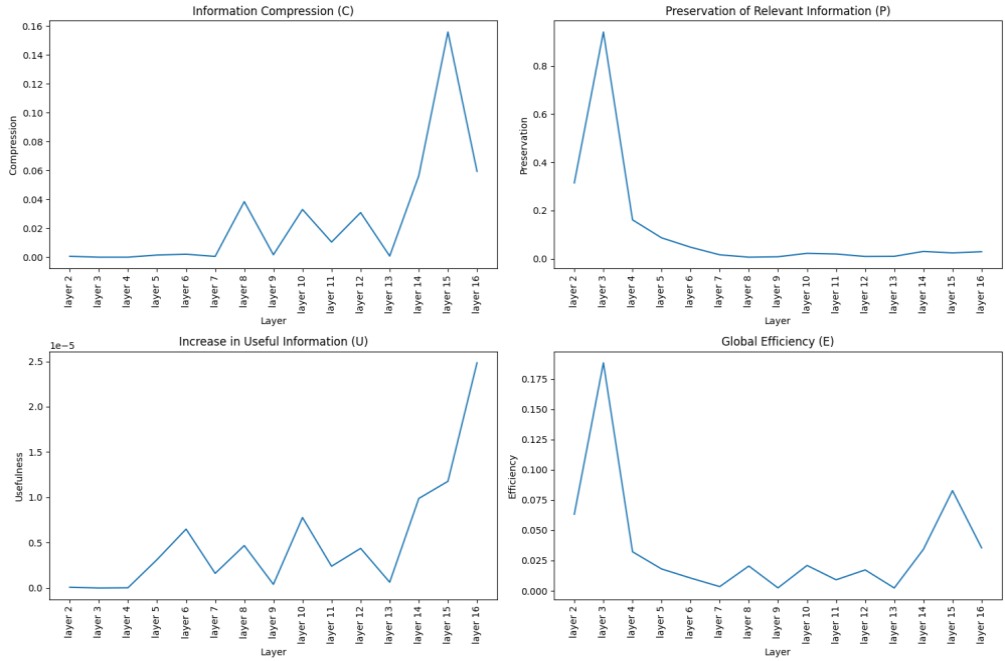

**Fig. 2:** Information Transformation Efficiency for ResNet50 on ImageNet

**b. Activation-Based Information Plane Analysis:**

For both VGG16 and ResNet50 (as shown in Figures 3 and 5), $I(T_l ; Y)$ is remarkably low for layers 1 to 3, and reaches its maximum at layers 7,8 and 10 for VGG16 and at layers 14 to 16 for ResNet50. For the two networks $I(X ; T_l)$ is the lowest at layer 4. Additionally, the standard deviation of $I(X ; T_l)$ is high for layers 1 to 3 and becomes comparatively very small from layers 6 to 15. The standard deviation of $I(T_l ; Y)$ remains of the same level for all the layers of the network.

Figures 3 and 5 also reveal a grouping of layers. For ResNet5, the Early Layers (1 to 3), Intermediate Layers (4 to 10) and Deep Layers (11 to 16). This grouping suggests different roles for these layers.

The activation-based information plane gives us insight on layers static states but does not give us a precise insight of the layer contribution to the information dynamics within the network. Some conclusions from the work of Tishby and Shwartz-Ziv (Shwartz-Ziv,2017) are somehow misleading, particularly regarding the attribution of fitting and compression functions to the layers at the end of the network. Additionally, the theoretical expectation of a monotonic decrease in both $I(X ; T_l)$ and $I(T_l ; Y)$ with increasing layer depth does not always hold true in practice. These observations will be discussed in more detail in a dedicated section later.

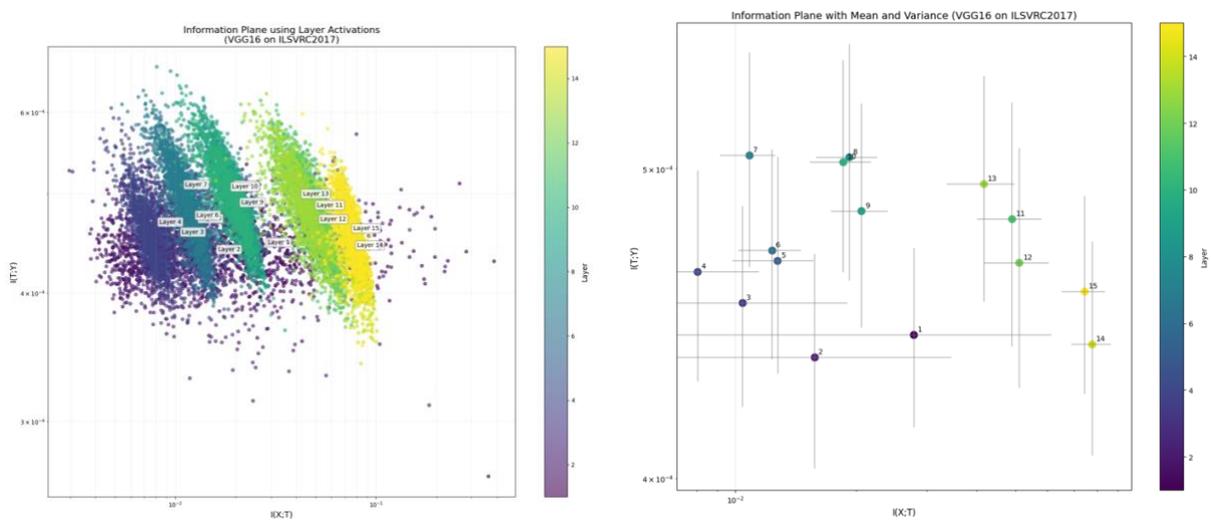

**Fig. 3:** Activation-Based Information Plane for VGG16 on ImageNet

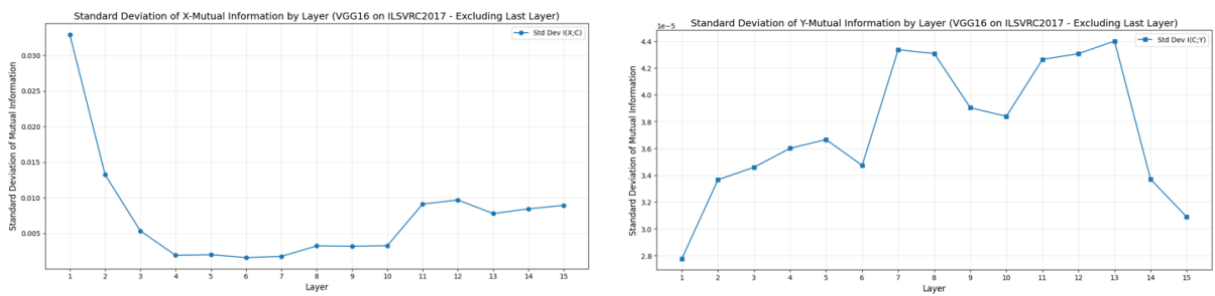

**Fig. 4:** Activation-Based Information Plane for VGG16 on ImageNet

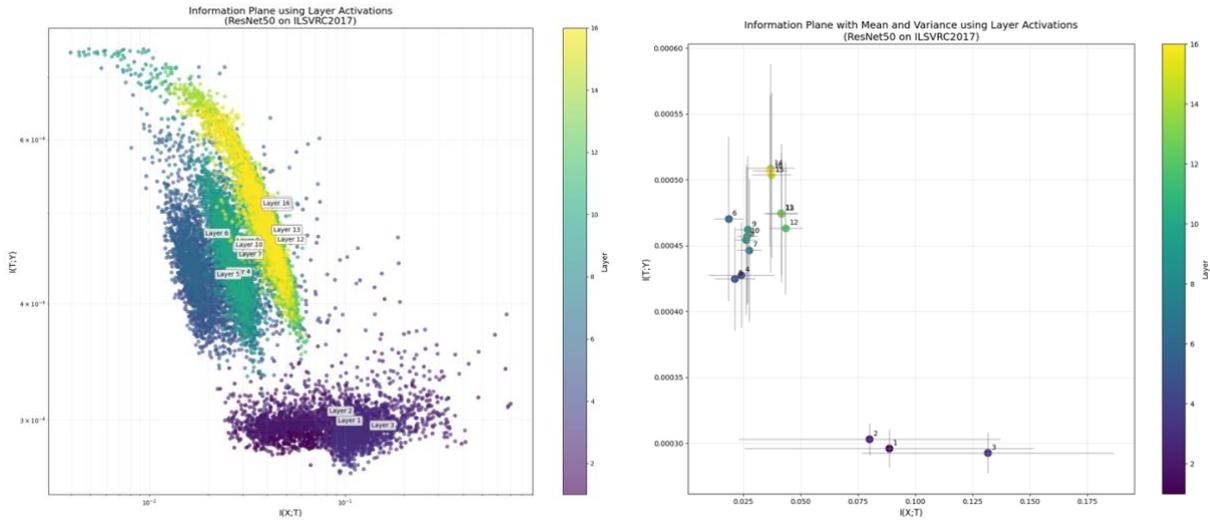

**Fig. 5:** Activation-Based Information Plane for ResNet50 on ImageNet

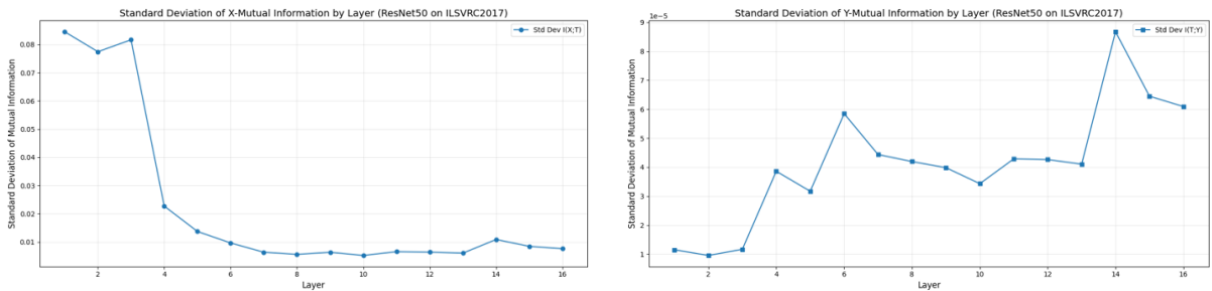

**Fig. 6:** Standard Deviation of X and Y-Mutual Information for ResNet50 on ImageNet

c. **Conductance-Based Information Plane Analysis:**

The use of layer conductance enables us to interpret the conductance-based information plane (Fig. 7 and Fig. 9) as a visualization of how each layer contributes to the overall information dynamics within the network. For any given layer $l$, $I(X\ ;\ C_l)$ characterizes its role in preserving or discovering relevant features from the input $X$, while $I(C_l\ ;\ Y)$ reflects its contribution to the flow of information that ultimately leads to predicting the output $Y$.

Focusing on ResNet50 (Figure 9), we observe a clear grouping of layers, each with distinct roles in information processing:

**Early Layers (2 to 6, beyond 3):** These layers primarily focus on orienting the information towards the expected output $Y$, as evidenced by the high $I(C_l\ ;\ Y)$, which reaches its maximum at layer 6 across the entire network. Additionally, the moderate $I(X; C_l)$ values suggest a "deconstruction" process of the input $X$, indicating these layers do not focus on preserving the input features. . This trend of non-preservation is also reflected in the ITE analysis (Figure 2).

**Intermediate Layers (7 to 10):** These layers strike a balance between orienting information towards $Y$ and compressing it. They effectively filtering out irrelevant information while maintaining task-relevant features. They effectively filter out irrelevant information while maintaining features essential for the task. This balance is reflected in both the ITE and conductance-based planes, showcasing their dual role in compression and preservation.

**Deep Layers (11 to 16):** These layers emphasize data compression and reconstruction of features relevant to the input $X$, as well as consolidating information pertinent to $Y$. Figure 9 shows a significant increase in $I(X; C_l)$ in these layers, indicating their role in reconstructing input features. Concurrently, there is a moderate increase in $I(C_l; Y)$, reflecting their contribution to the output prediction.

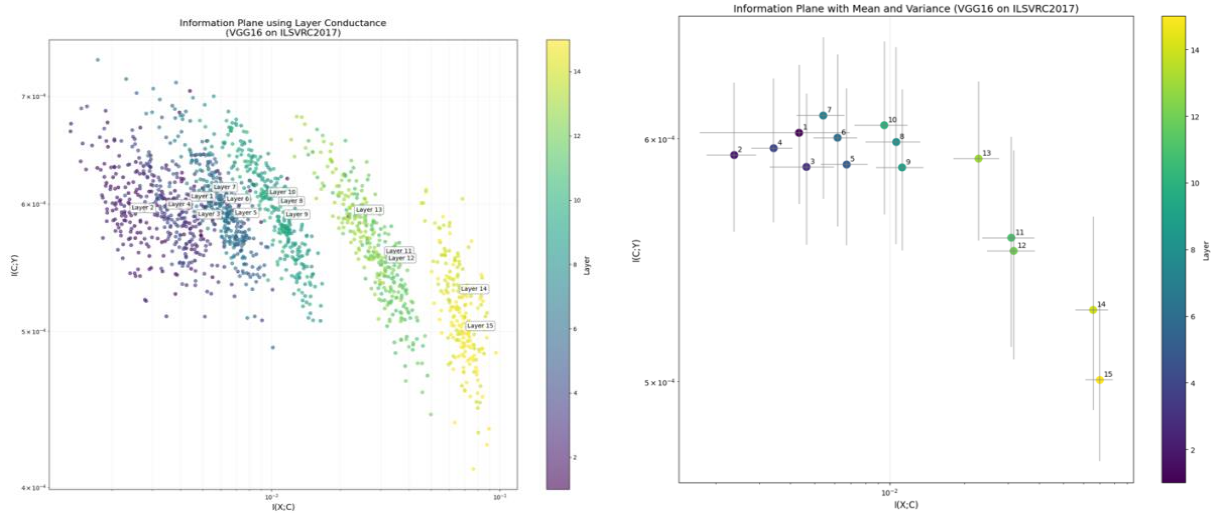

**Fig. 7:** Conductance-Based Information Plane for VGG16 on ImageNet

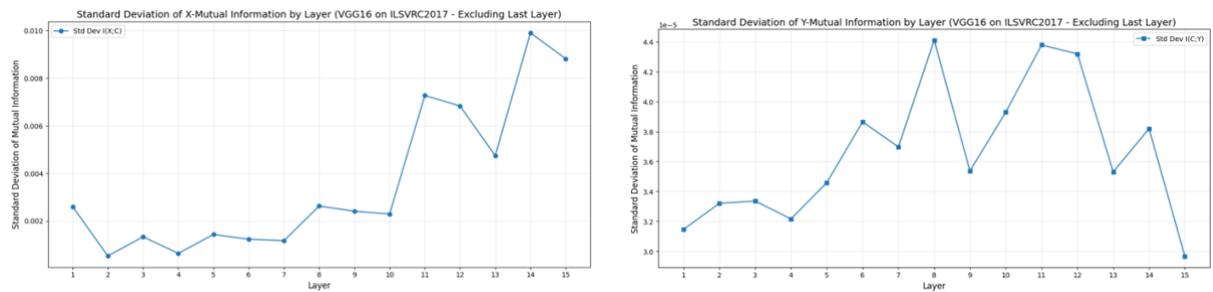

**Fig. 8:** Standard Deviation of X and Y-Mutual Information for VGG16 on ImageNet

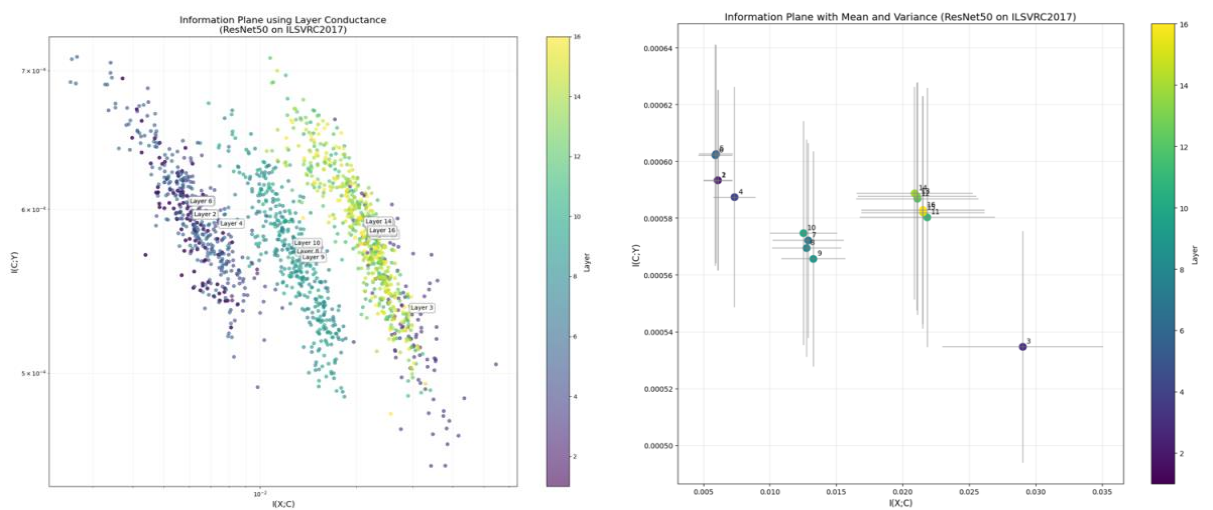

**Fig. 9:** Conductance-Based Information Plane for ResNet50 on ImageNet

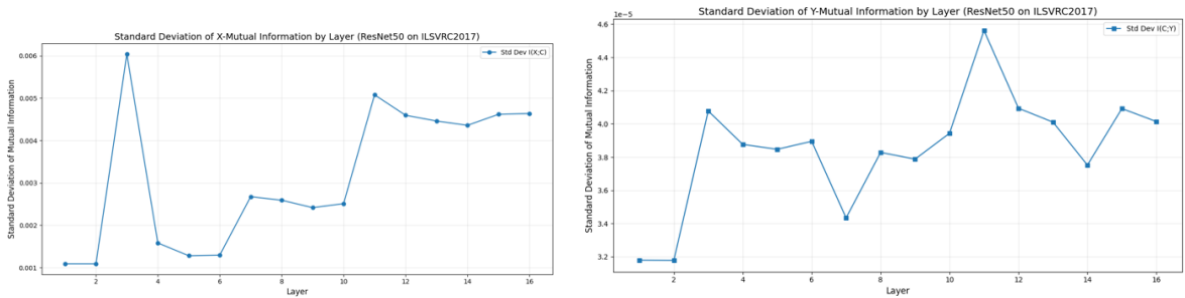

**Fig. 10:** Standard Deviation of X and Y-Mutual Information for ResNet50 on ImageNet

**d. Consistency Across Metrics:**

The consistency between the ITE and the conductance-based information plane reinforces the validity of our approach. Both metrics highlight the critical layers and their roles in the information flow within the network. The granular insights provided by conductance help in understanding the specific contributions of each layer, which is less apparent from activation-based analysis alone.

This layered analysis, facilitated by the Conductance-based Information Plane, provides an understanding of how different stages of the network contribute to information processing. The early layers act as initial filters, the middle layers refine and compress, and the later layers focus on extracting the most salient features in the input, in order to achieve the prediction.

In the ITE-Preservation graph (Figure 2), Layer 3 shows a significant contribution to preserving relevant information. This aligns with the conductance-based information plane (Figure 9), where Layer 3 exhibits high mutual information value $I(X; C_l)$, indicating its critical role in maintaining information fidelity.

**e. Empirical Contradictions with Information Bottleneck Theory**

In the work by (Tishby et al. 2015), it is argued that each layer in a deep neural network processes inputs solely from the preceding layer, forming a Markov chain:

$$Y \rightarrow X \rightarrow A_1 \rightarrow \ldots \rightarrow A_l \rightarrow \ldots \rightarrow A_L = \hat{Y}$$

Here $A_l$ denotes the $l$-th hidden layer activations and $A_L = \hat{Y}$ denotes the output of the network. And then we get the Data Processing Inequality (DPI):

$$I(X;Y) \geq I(A_1;Y) \geq I(A_2;Y) \geq \ldots \geq I(\hat{Y};Y)$$

$$I(X;X) \geq I(X;A_1) \geq I(X;A_2) \geq \ldots \geq I(X;\hat{Y})$$

These inequalities imply that information about $X$ or $Y$ that is lost in one layer cannot be recovered in subsequent layers. However, empirical observations using the ResNet50 pre-trained network on the ImageNet dataset, as well as other networks on different datasets, reveal that mutual information can sometimes increase in higher layers. This results in instances where $l < k$, and $(X;A_k) \geq I(X;A_l)$ or $I(A_k;Y) \geq I(A_l;Y)$.

Importantly, this observation persists across a comprehensive range of mutual information estimation techniques, including binning methods, kernel density estimation, k-nearest neighbor approaches, and neural estimation techniques. The consistency of results across diverse estimation methods strongly suggests that the observed contradiction is not an artifact of the estimation process but an inherent phenomenon in DNNs. Some possible explanations are:

- DPI may hold on average but not for specific input subsets, particularly those used in training and similar real-world data.
- DNNs optimization for training data, potentially violating general theoretical constraints.
- Intermediate layers might expand the feature space, temporarily increasing mutual information.
- Complex non-linearities can reorganize information, apparently increasing mutual information.

It is crucial to note that the set of "natural" inputs (i.e., data used for training and evaluation and real-world data) represents a subset of measure zero within the space of all possible inputs (e.g., random images). Therefore, the observed properties of mutual information on this subset may not generalize to all possible inputs. This nuanced understanding highlights the importance of considering both theoretical principles and empirical observations when analyzing the information flow in DNNs.

This insight underscores the complexity of information dynamics in DNNs and the need for further research to reconcile theoretical principles with practical implementations.

## 4. Conclusion

This study has highlighted the critical importance of analyzing the flow of information through the hidden layers of neural networks using the framework of the Information Plane and a conductance-based approach. Our empirical results, derived from pre-trained models such as ResNet50 and VGG16 on the ImageNet dataset, demonstrate that certain hidden layers play a pivotal role in capturing and preserving essential features, thereby enhancing not only model performance but also interpretability.

We found that intermediate layers effectively balance information orientation and compression, filtering out irrelevant information while retaining pertinent features for the task at hand. Furthermore, our approach revealed that deeper layers focus on compressing data and optimizing relevant information, underscoring the complexity of information dynamics in deep neural networks.

The implications of this research are significant. On one hand, it provides more precise analytical tools for diagnosing issues such as overfitting. On the other hand, it paves the way for the design of more robust and efficient models capable of better generalization on unseen data. By integrating measures of information transformation and considering the dynamics of information flow, we can gain a deeper understanding and explanation of neural network functioning, which is essential for their adoption in critical applications.

Hence, while our study has revealed complexities and deviations from established theoretical principles, it underscores the need for ongoing research to reconcile empirical observations with the theoretical foundations of deep learning. The future of model interpretability in neural networks will depend on our ability to navigate these two dimensions, leveraging the insights provided by approaches such as the one we have proposed.